\documentclass[a4paper,fleqn]{cas-dc}

\usepackage[numbers]{natbib}
\usepackage{emptypage}

\def\tsc#1{\csdef{#1}{\textsc{\lowercase{#1}}\xspace}}
\tsc{WGM}
\tsc{QE}
\tsc{EP}
\tsc{PMS}
\tsc{BEC}
\tsc{DE}

\begin{document}
\let\WriteBookmarks\relax
\def\floatpagepagefraction{1}
\def\textpagefraction{.001}
\shorttitle{Cross-domain representation}
\shortauthors{Prasad and Singh}

\title[mode=title]{Cross-Domain Identity Representation for Skull to Face Matching with Benchmark DataSet}

\author[1]{Ravi Shankar Prasad}[orcid=0009-0003-7767-6762]
\ead{d23033@students.iitmandi.ac.in}

\author[1]{Dinesh Singh}
\ead{dineshsingh@iitmandi.ac.in}

\affiliation[1]{
  organization={Visual Intelligence and Machine Learning (VIML) Group, 
  School of Computing and Electrical Engineering, Indian Institute of Technology Mandi},
  country={India}
}

\cortext[cor1]{Corresponding author}

\begin{abstract}
Craniofacial reconstruction in forensic science is crucial for the identification of the victims of crimes and disasters. The objective is to map a given skull to its corresponding face in a corpus of faces with known identities using recent advancements in computer vision, such as deep learning. In this paper, we presented a framework for the identification of a person given the X-ray image of a skull using convolutional Siamese networks for cross-domain identity representation. Siamese networks are twin networks that share the same architecture and can be trained to discover a feature space where nearby observations that are similar are grouped and dissimilar observations are moved apart. To do this, the network is exposed to two sets of comparable and different data. The Euclidean distance is then minimized between similar pairs and maximized between dissimilar ones. Since getting pairs of skull and face images are difficult, we prepared our own dataset of 40 volunteers whose front and side skull X-ray images and optical face images were collected. Experiments were conducted on the collected cross-domain dataset to train and validate the Siamese networks. The experimental results provide satisfactory results on the identification of a person from the given skull.


\end{abstract}




\maketitle

\section{Introduction}

Craniofacial recognition aims to identify a person based on the underlying skull~\cite{damas2011forensic}. Craniofacial recognition is used in various applications such as forensics, history, anthropology, and archaeology~\cite{damas2020relationships}.
    In forensic science, facial recognition is particularly useful for identifying deceased individuals when all other means of identification have been exhausted but skeletal remains~\cite{missal2023forensic}. When an unidentified skull is discovered at a crime scene, law enforcement officials first reconstruct the face by overlying clay on the skull and then compare it with a database of facial images of missing persons to establish its identity. Identification of terrorists is one of the many forensic cases that benefit significantly from successful identification, including the identification of victims of large landslides in mountainous regions and during tsunami in coastal regions. Soft tissues like skin, hair, and other organs starts decomposing over time period, but skeletons are available for a longer time period that can be used to identify the person.

\begin{figure}[!t]
        \centering
        \centering
        \includegraphics[width=1.5\linewidth, keepaspectratio, trim={4.8cm 6cm -2.4cm 5.5cm},clip]{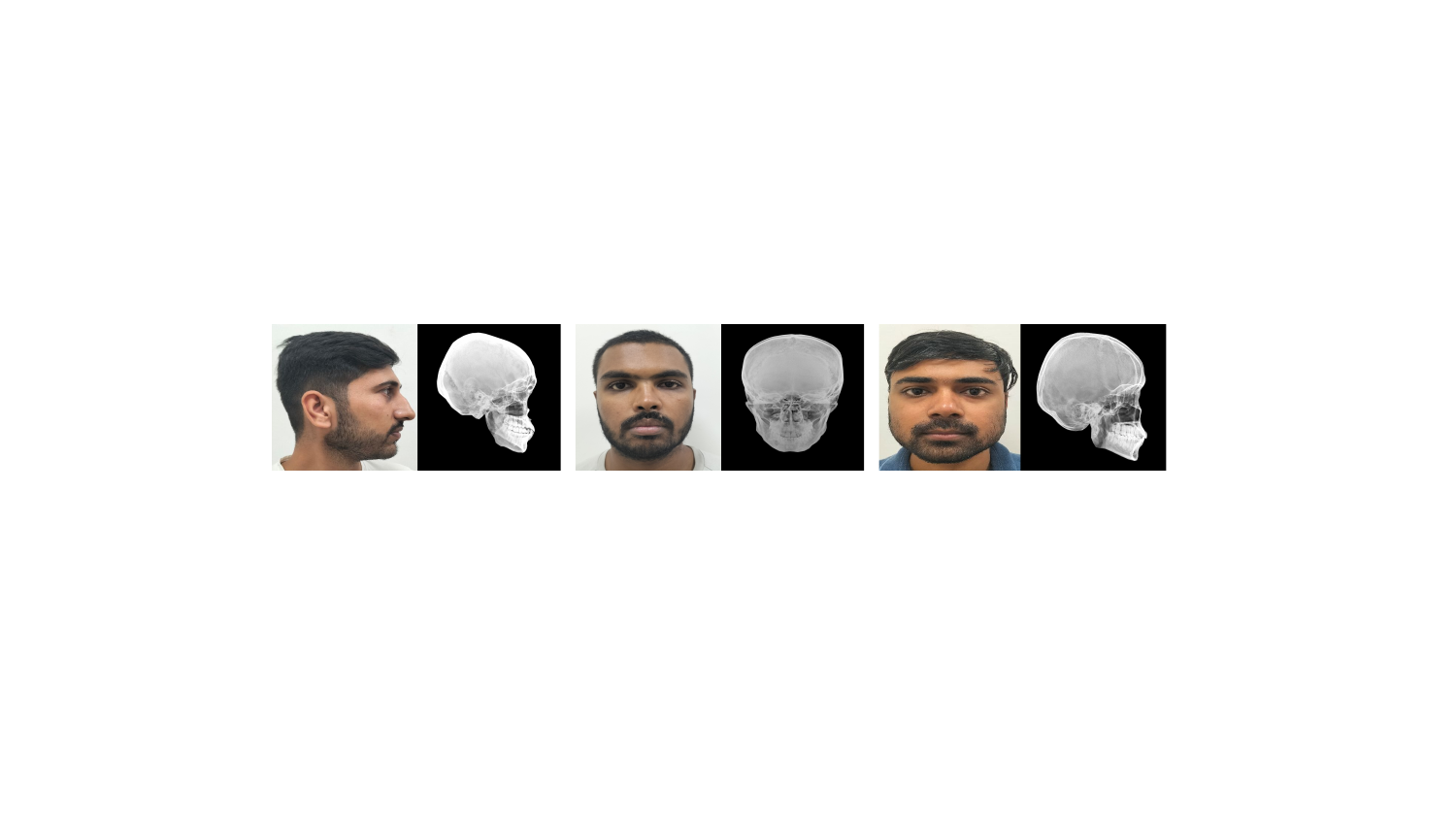}
        \text{\small(a) Positive image pairs}
        \hfill
        \centering
        \includegraphics[width=1.5
        \linewidth, keepaspectratio, trim={4.8cm 6cm -2.4cm 5cm},clip]{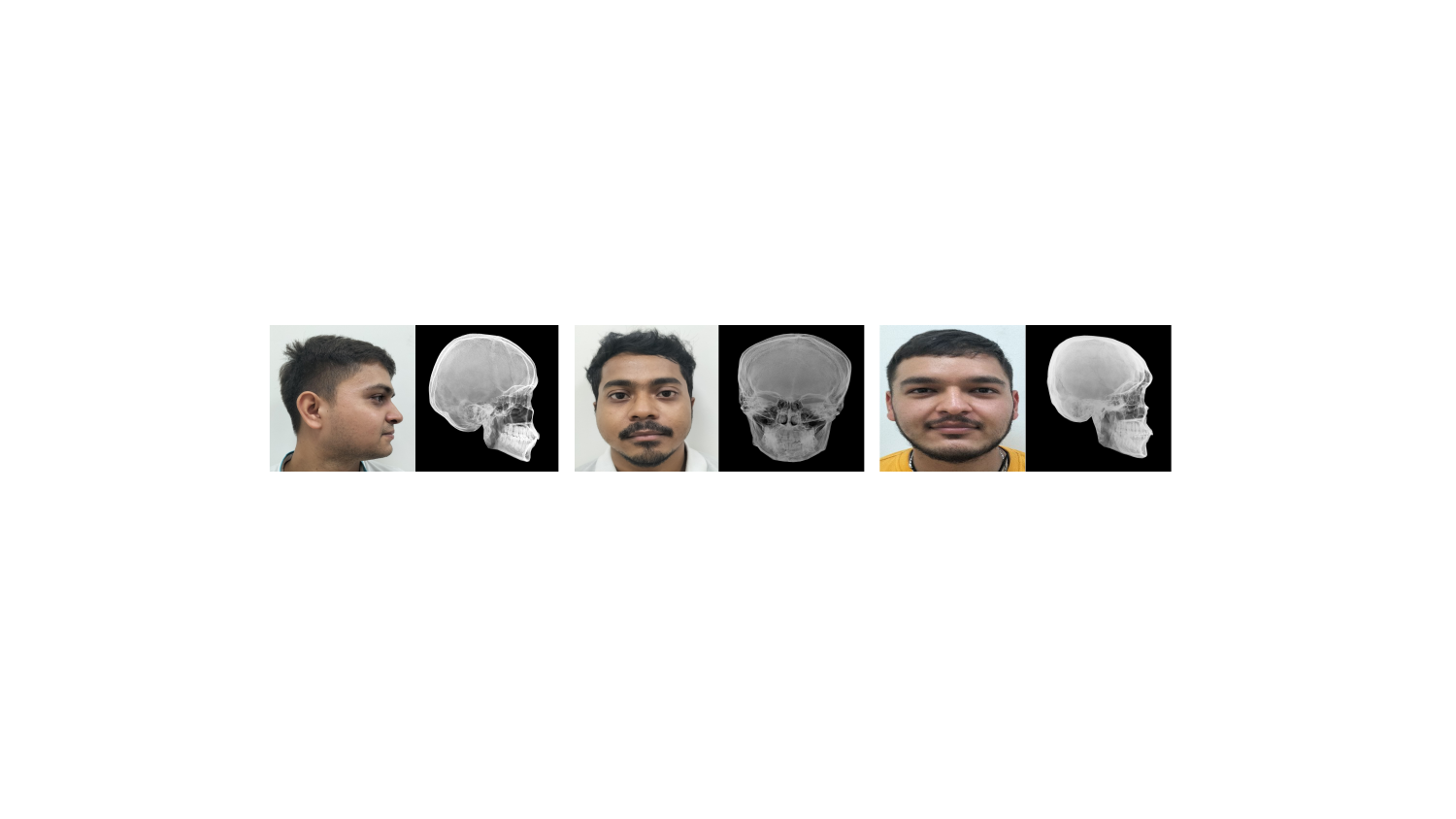}
        \text{\small(b) Negative image pairs}
        \caption{Randomly picked positive and negative pairs for evaluation dataset.}
        \label{fig:image_pairs}
    \end{figure}

 \begin{figure*}[!ht]
        \centering
        \includegraphics[width=1\linewidth, keepaspectratio, trim={0cm 8cm -0.4cm 7.5cm},clip]{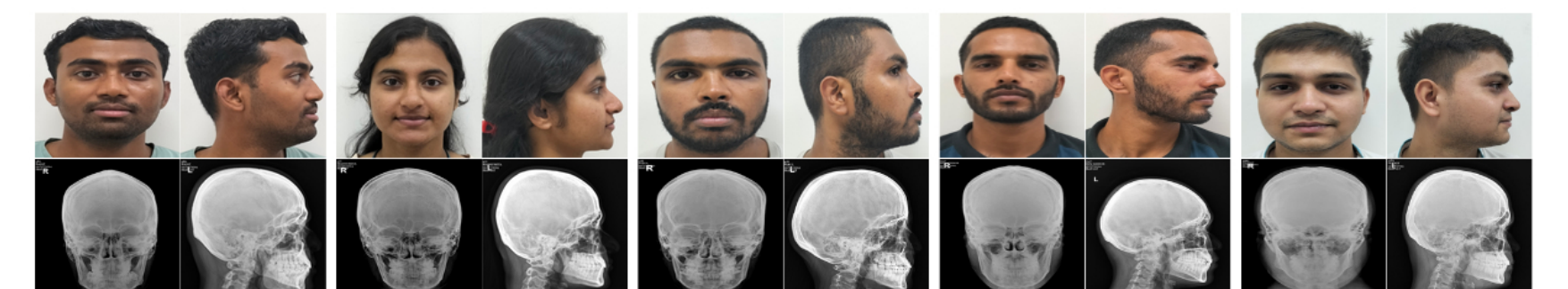}
        \caption{Paired face and X-ray samples from the prepared dataset \emph{{IITMandi\_S2S}}
        for skull to face matching}
        \label{fig:dataset}
    \end{figure*}
    With the advancement of ubiquitous imaging technologies such as optical cameras and X-ray imaging, capturing faces and skulls has become more affordable and accessible. Also, non-intrusive 3D face data capturing techniques can be used~\cite{xie2023mm3dface}. More crucially, 3D face modeling from photographs has received a lot of attention~\cite{jin2017robust}. When applying template deformation techniques, it is assumed that if the skulls are comparable in any way, then the associated face\textquotesingle s primary characteristics should share shape~\cite{muller2005template}. In general, certain visually unexpected aspects of the template face may appear in the reconstruction results produced by these kinds of approaches. Traditional methods involve shaping a moldable material by hand onto an unidentified skull using anatomical data and reference material. According to Wilkinson\textquotesingle s detailed examination, this is a very subjective process that requires a great deal of creative interpretation and sometimes produces random results~\cite{wilkinson2010facial}. Later on, mainly there are two main methods for this process. The first method involves craniofacial superimposition to directly compares the 2D skull image with the face image~\cite{damas2011forensic}. The second method aims to construct a 3D face model using craniofacial reconstruction technology and then compare this model with facial images of missing persons~\cite{claes2010computerized}. Tu~\textit{et al.} proposed a landmark-based recognition system for skull identification~\cite{tu2007automatic}. The method involves identifying the face by extracting landmarks from both the 3D and 2D faces and then determining the reprojection errors. While this approach is user-friendly and feasible, it does not account for the varying quality of landmarks on the face. In the improvement of skull identification using a landmark-based algorithm, Huang~\emph{et al.} proposed a new landmark-based algorithm that is based on the quality analysis of landmarks~\cite{huang2011weighted}, where different weights of the landmarks have been taken into account in recognition. But, again, this method fails to get the optimal landmark points on the face as landmarks on the 2D and 3D faces are done manually. All these methods face challenges with data insufficiency due to the non-availability of enough pairwise data of the skull and its respective face image. 

    In this study, we first created our benchmark dataset \emph{IITMandi\_S2F}, which includes an X-ray image and its corresponding frontal and side face images. Then, we utilized a deep neural network, specifically the Siamese networks ~\cite{melekhov2016siamese}, to match a person\textquotesingle s skull in a given set of faces. This was achieved by computing a similarity metric based on the Euclidean distance between the feature representations on each side of the network and training the model with a triplet loss function. We evaluated the performance of the trained model on our benchmark dataset for skull-to-face identification. The collected benchmark dataset can also be used for other related research, such as craniofacial superimposition and craniofacial reconstruction. The overall contribution of this research work are as follows
    \begin{itemize}
        \item Prepared an benchmark dataset \emph{IITMandi\_SF} to kickstart the research on craniofacial recognition and reconstruction. The dataset consists of paired X-ray images of skulls and optical images of people's faces taken from both front and side angles, collected from 40 volunteers.
        \item A novel framework for learning cross-domain identity representation using Siamese networks.
        \item Performance evaluation of the deep neural networks, such as Siamese networks on this dataset for matching a skull to a corresponding face.
    \end{itemize}

    \section{\textbf{Related Work}}
    Due to the unavailability of the public benchmark dataset, limited research is carried out for the automation of the person identification based on the skull. At the moment, artists or anatomists perform craniofacial reconstruction by hand. To address actual 3D shape issues, several surface-based shape analysis and classification techniques were also put forth\cite{paysan2009face}. A few automatic techniques for skull-to-face overlays have recently been put forth. Through the use of the link between soft tissues and the underlying bone structure, craniofacial reconstruction~\cite{claes2010computerized} attempts to approximate an individual's face appearance from its skull. However with the limited amount of pairwise image datasets of skulls with known facial images, this approximation is not well established. Recent methods use statistical learning techniques to explore the relationship between the skull and face, such as statistical shape models~\cite{claes2010bayesian, hu2013hierarchical}, regression models~\cite{berar2011craniofacial, duan2014craniofacial}, etc. But the statistical models do not capture individual variation, thus most reconstructed faces are not good enough for identification. They merely represent craniofacial variation statistically.

    Both craniofacial superimposition and reconstruction methods need to have a strong comprehension of the anatomical relationship between the skull and the face. When identifying an unknown skull, the reconstruction process is time-consuming and subjective. A few computerized reconstruction methods~\cite{vandermeulen2006computerized, michael19963d} deformed a craniofacial reference to obtain the face of the unknown target skull. This was done by deforming the reference skull to the target skull based on certain skull features and then extrapolating the skull deformation to the reference face in order to obtain the reconstructed face. These methods assumed that the reference and the unknown target individual had similar tissue thickness distributions. However, using an inappropriate reference could lead to biased or unrealistic reconstructions~\cite{claes2010computerized}, but an accurate and robust representation of the relationship between the skull and face is still intractable and unsolved.

    In this paper, in order to have a strong comprehension of the anatomical relationship between the skull and face, we propose to identify an unknown skull by using a Siamese network between the 2D skull and 2D face in terms of the morphology, and measure the the similarity score between the unknown skull and the faces.

    \begin{figure*}[!ht]
        \centering
               \includegraphics[width=1.0\linewidth, keepaspectratio, trim={0.5cm 3cm 0.5cm 6cm},clip]{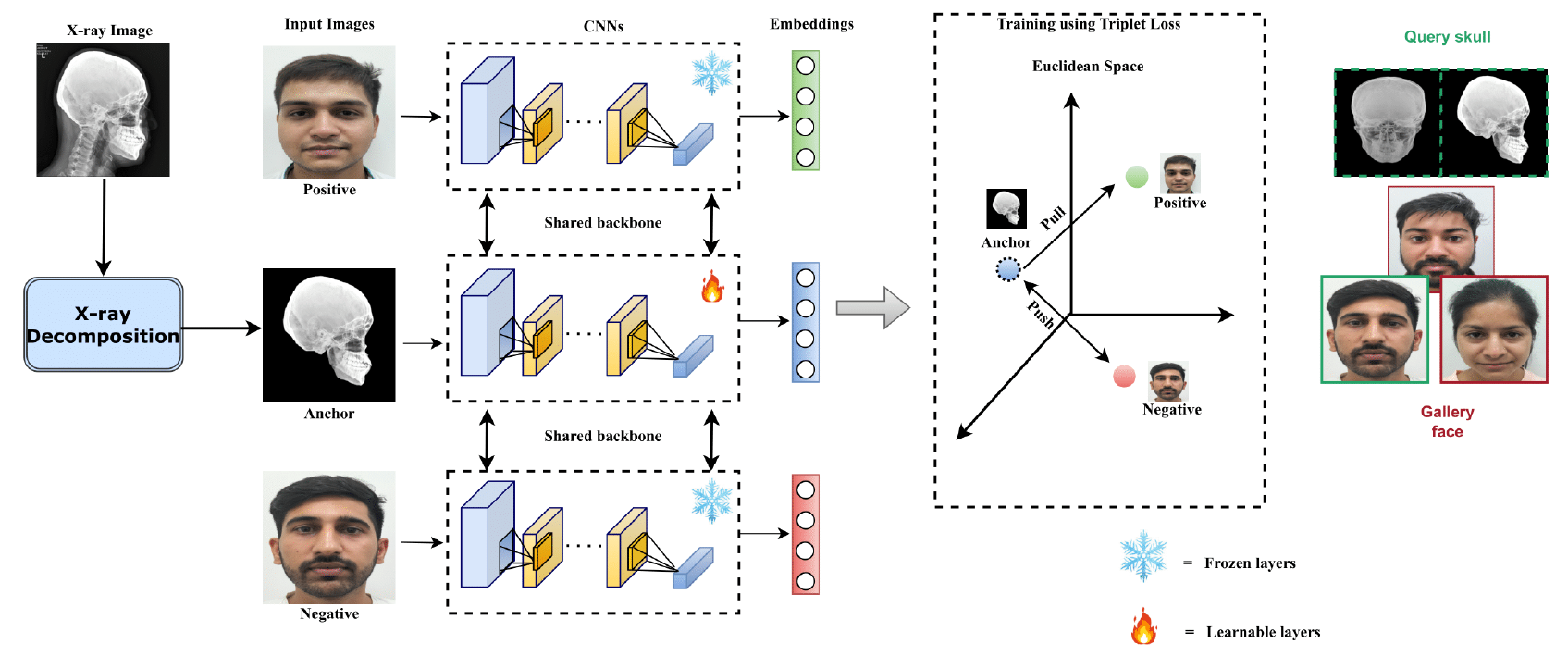}
        \caption{Proposed Overall Framework for Skull-to-Face Image Retrieval and Architecture Using Triplet Loss. In this framework, Convolutional Neural Networks (CNNs) serve as the backbone of our proposed Siamese networks, which share a common backbone. For the face images, we utilize pretrained deep models with frozen layers, while for the skull images, we train the same pretrained model with adjustable weights. On the right side, we provide an illustration of the retrieval process, showing how a face is retrieved from the face gallery using a query skull image. }
        \label{fig:siamese}
    \end{figure*}

\section{\textbf{Details of the dataset}}\label{sec:dataset}
    Deep neural networks require massive datasets for generalized model training, while getting paired data of face and skull is almost impossible at a large scale. However, to overcome this issue, we have used a face X-ray scan of a person from both front and side views. X-ray, an affordable and ubiquitous imaging technique, provides significant information on skull structure and can be conducted for all cases, such as skeletons, degraded bodies, and living persons. Fig.~\ref{fig:image_pairs} shows some samples of pairwise dataset. In this study, we prepared pairwise datasets where each subject's optical and X-ray images of faces are collected with the front and side views as shown in Fig.~\ref{fig:dataset}. First, we decompose the X-ray of a person by removing the soft tissue part from its X-ray image, and the remaining hard tissue image of the X-ray is kept so that it resembles the skull of that person, which is shown in Fig.~\ref{fig:siamese}. Our study was conducted on the database of 40 X-ray scans with their respective face pair image from volunteers who mostly came from the North and East regions of India, aged 21-30 years females and males. In this study, there were 18 females and 22 number of males. The X-ray image was obtained in India by a clinical X-ray scanner system from the health center at IIT Mandi, Himachal Pradesh. The optical images of each subject are taken by the 50 megapixel camera. The X-ray and optical pairwise images of each subject were stored in JPG format with size ranges from  $1000\times 2000$ to $2000\times 3000$ pixels. 

For each volunteer, $4~(=2 \times 2)$ anchor-positive pairs are created, while each anchor-positive pair can have $78~(=39 \times 2)$ negatives. Thus, for $n=40$ volunteers, a total of $12,480~(=4 \times 78 \times n)$ triplets are formed. Finally, we split the triplets into training and validation sets with a $70:30$ ratio, respectively.

\section{\textbf{Cross-Domain Identity Representation}}
   We used the Siamese networks for cross-domain identity representation, calculating the matching score between the provided skull and face images. Siamese neural networks consist of twin networks that share weights~\cite{melekhov2016siamese} and are generally trained using triplet loss. However, in our approach, we employed pretrained deep models as a backbone of Siamese networks to extract the embeddings from the face images. Since the skull images come from a different domain, we maintained the same backbone architecture, but incorporated learnable layers to train the embeddings for the skull images.

The loss function evaluates the deviation in the learned representation of the two input images, using an anchor image~\cite{shabanov2024stir}. In this context, the two input images are a positive and a negative face image, while the anchor is a skull image. We utilized a similarity metric based on Euclidean distance. The triplet loss function is defined as follows:

\begin{equation}\label{eq:loss}
L_t = \frac{1}{N} \sum_{i=1}^{N} \max(0, d(f(x^a_i)^*, f(x^p_i)) - d(f(x^a_i)^*, f(x^n_i)) + \alpha)
\end{equation}

Where \( N \) represents the total number of triplets in a batch. Each triplet consists of three images: an anchor image \( x^a_i \), a positive image \( x^p_i \) (which belongs to the same class as the anchor), and a negative image \( x^n_i \) (which belongs to a different class). The function \( f(x) \) is a backbone neural network used in proposed Siamese networks, that maps an input image to a d-dimensional feature space, where \( f(x^p_i) \), \( f(x^n_i) \) denote the embeddings of the positive, and negative images respectively and  \( f(x^a_i)^* \) is the embeddings of anchor. \(f()^*\) symbols represent learnable parameters.

Here, $\alpha$ is an important hyperparameter that establishes the minimum permissible difference between the distances of anchor-positive pairs and anchor-negative pairs. By defining this threshold, one ensures that the model effectively differentiates between similar and dissimilar data points, thereby enhancing its overall performance and accuracy. The function \( \max(0, \cdot) \) ensures that only triplets violating the margin constraint contribute to the loss, meaning the loss is nonzero only when the negative image is not sufficiently far from the anchor compared to the positive image.

     The Siamese networks seek to push feature vectors away from input pairs that are labeled as dissimilar and to bring the output feature vectors closer to input pairings that are labeled as similar, in contrast to traditional techniques that provide binary similarity labels to pairs. Also, the objective of triplet learning is to learn meaningful representations by comparing anchors with positive and negative samples in the dataset using a distance metric.~\eqref{eq:loss} helps to organize the latent space into similar and dissimilar embedding\cite{bachman2019learning}, \cite{hadsell2006dimensionality} by using anchor and pair embeddings that are positive and negative. In order to force the networks to learn the cross-domain identity representation, we train on positive and negative pairs formed across the X-ray and optical images. Fig.~\ref{fig:siamese} depicts the overall architecture of Siamese networks, which consists of three backbone networks that share a common structure. For face images, we employ pretrained deep models with their layers frozen, while for skull images, we train the same pretrained model with learnable weights. Training is performed using triplet loss, as outlined in equation (1). In this scenario, the anchor image represents the skull image of an individual, which is obtained by extracting the hard tissue from its X-ray image, thus resembling the skull. Both the anchor and positive images share the same ID, whereas the negative image is assigned a randomly selected ID that differs from that of the anchor.

     \begin{table*}[t]
    \centering
     \caption{Model performance with different hyperparameter configurations on our benchmark dataset. Recall, mAP and MRR are for the top 30 with a gallery containing 40 face images. Models with the best performance are marked in bold. }
    \begin{tabular}{|c|c|c|c|c|c|}
        \hline
        \textbf{Model}  & \textbf{Training accuracy} & \textbf{Validation accuracy} & \textbf{Recall@30} & \textbf{mAP@30}& \textbf{MRR@30}  \\
        \hline
        VGG16    &0.6562 & 0.6250 & 0.8692 &0.5031 &0.4259\\
        \hline
      {ResNet18}   & 0.7812  & 0.6250 &  0.8389& 0.5551  &0.8333 \\
        \hline
        \textbf{ResNet50}  & \textbf{0.8750}  & \textbf{0.7500} &   {0.8692}& {0.6932}&0.8333 \\
        \hline
        ResNet101& 0.6562& 0.6250 &  0.8692&0.7245 &1.0000 \\
        \hline
        \textbf{DenseNet201} & \textbf{ 0.7812}  &\textbf{0.7500}  & {0.7783}& {0.5217} &0.4722  \\
        \hline
         EfficientNet\_B0  & 0.7188  & 0.6250 &  0.8692&0.5282 &0.4556 \\
        \hline
        \textbf{MobileNet\_v2}& \textbf{0.7812}& \textbf{0.7500} &   0.8086&0.5910&0.8333  \\ \hline
        
        ConvNeXt\_Tiny&  0.6875& 0.6250 &  0.6874&0.5664 & 0.5278 \\
        \hline
        
        ViT\_B\_16 & 0.6250 &0.5000& 0.7177 & 0.6267 &0.8333\\
        \hline
        
    \end{tabular}
    \label{Table 1}

\end{table*}

\begin{figure}[!t]
        \centering
        \begin{minipage}{0.49\linewidth}
        \centering
        \includegraphics[width=1\linewidth, keepaspectratio]{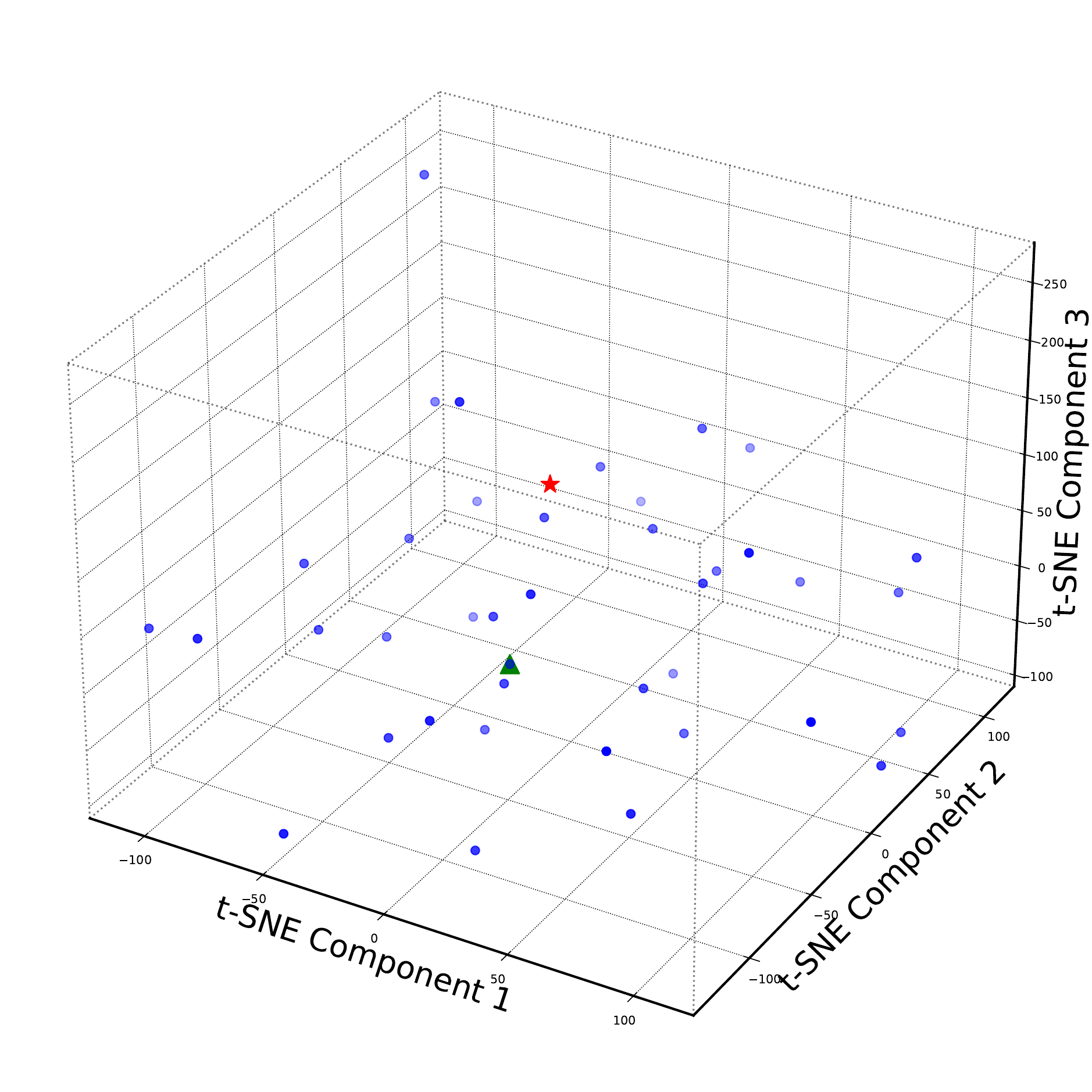}
        \end{minipage}
        \hfill
        \begin{minipage}{0.49\linewidth}
        \centering
        \includegraphics[width=1\linewidth,keepaspectratio]{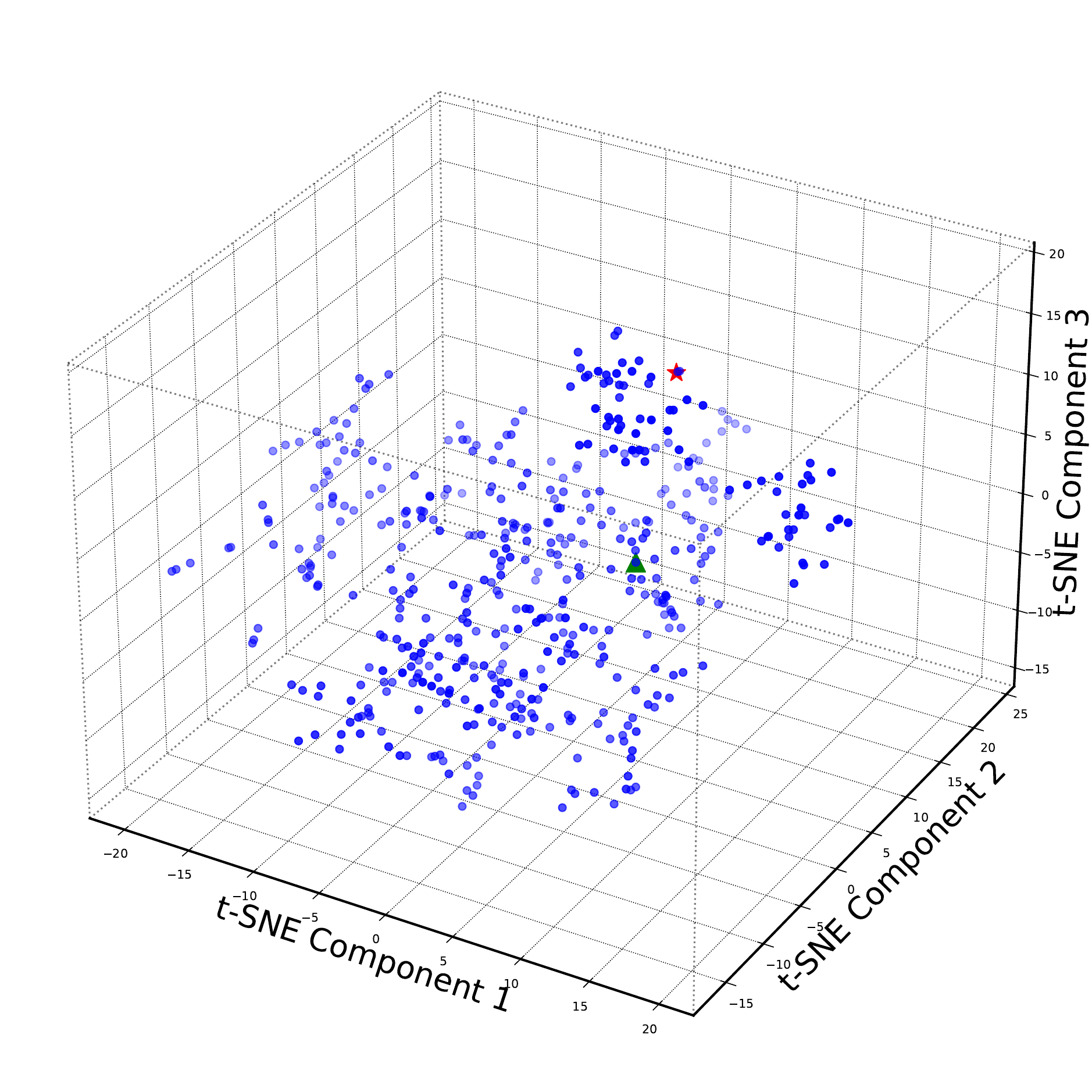}
        \end{minipage}
        \caption{3D t-SNE on the left side visualizes query skull embedding in red star, ground truth face embedding in green triangle, and \emph{IITMandi$\_$S2F} gallery face embeddings in blue circle. While 3D t-SNE on the right side visualizes the same, but with a mixed gallery of face embeddings, which are from \emph{IITMandi$\_$S2F} $+$ IISCIFD face dataset. (Best viewed in colors.) }    \label{fig:3d_tsne}
    \end{figure}
\begin{figure}[!ht]
        \centering
        \includegraphics[width=0.9\linewidth, keepaspectratio]{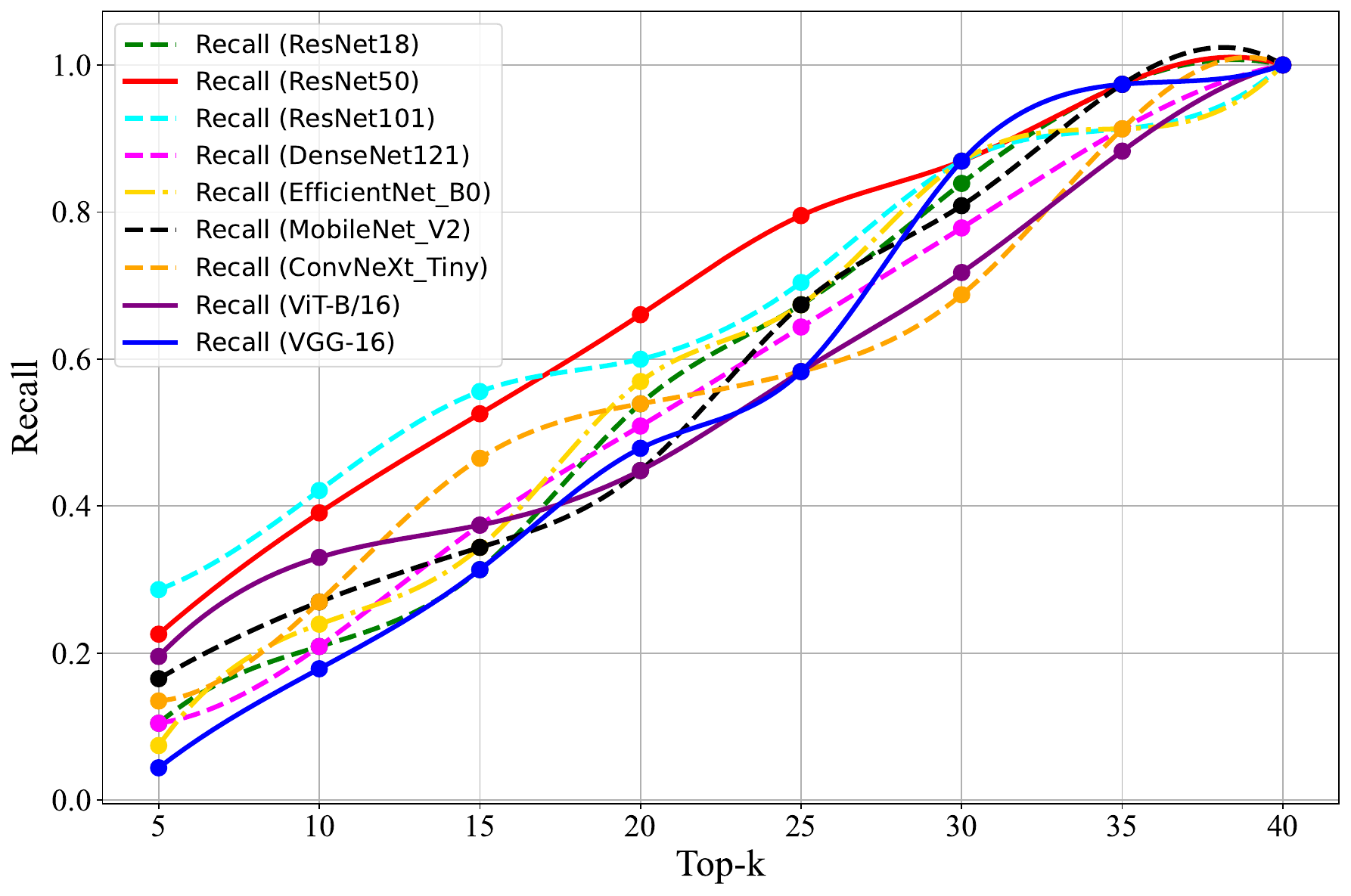}
        \caption{\textbf{Recall} results for retrieving the top k matched faces from the face gallery based on the confidence score based on the Euclidean distance between the given \textbf{query skull} and the gallery face using different models.}
        \label{fig:recall}
    \end{figure}

\begin{figure}[!ht]
        \centering
        \includegraphics[width=0.9\linewidth, keepaspectratio]{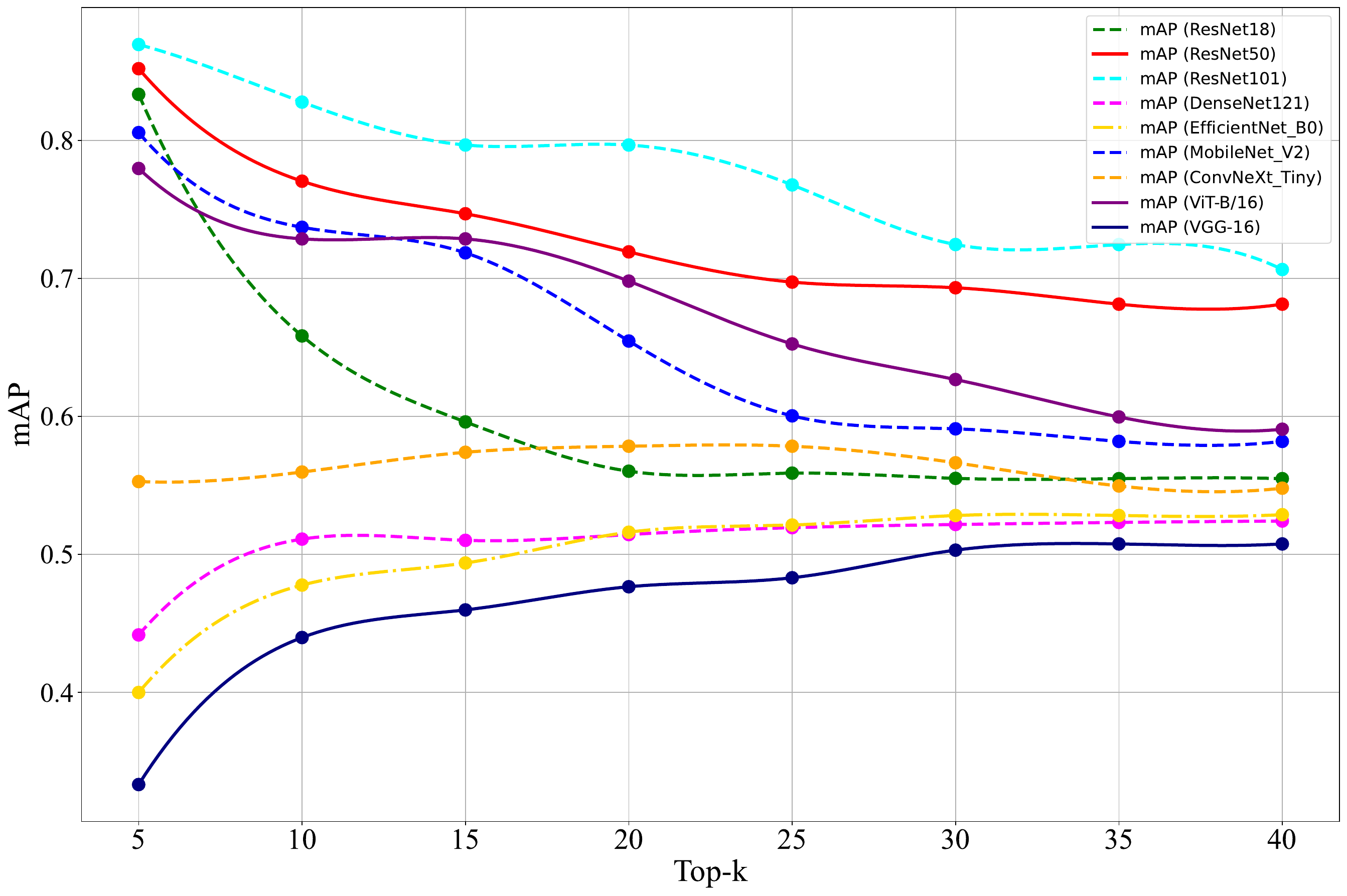}
        \caption{\ \textbf{Mean average precision} results for retrieving the top k-matched faces from the face gallery based on the confidence score based on the Euclidean distance between the given \textbf{query skull} and the gallery face using different models.}
         \label{fig:precision}
    \end{figure}
\section{\textbf{Experimental Evaluation}}
    A thorough validation of the developed automatic skull-to-face matching system is extremely crucial for the consideration of its practical use in real-life and to provide scientific evidence supporting the enhanced utility of this recognition technique for criminal investigations. The proposed skull-to-face similarity matching framework is evaluated on our prepared dataset. The details of the prepared dataset are provided in \S\ref{sec:dataset}. We use the $70\%$ of the image pairs for the training of the proposed model. During training, five types of data augmentation, like rotation, flipping, color jitter, random affine and brightness level changes of the skull and face images are applied to obtain a robust and generalized model that is invariant to rotation, reflectance, color, illumination, position and scale of the skull and face in the corresponding images.

\begin{figure}[!ht]
        \centering
        \includegraphics[width=0.9\linewidth, keepaspectratio]{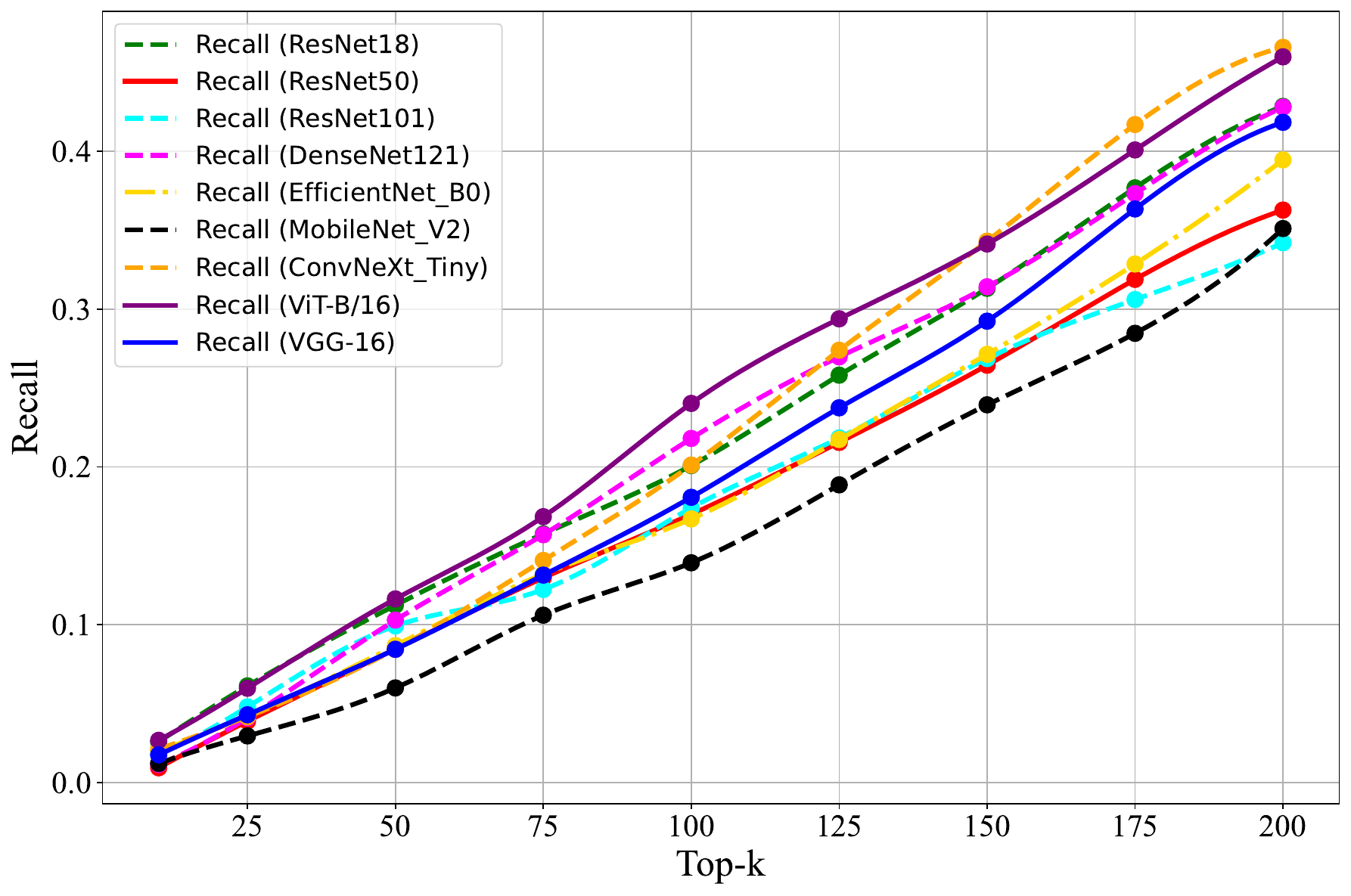}
        \caption{Comparison of various models on \textit{recall} results for retrieving the top $k$ matched faces given \textbf{query skull} from the \textbf{mixed gallery face} using different models.}
        \label{fig:recall_mixed}
    \end{figure}
    
\begin{figure}[!ht]
        \centering
        \includegraphics[width=0.9\linewidth, keepaspectratio]{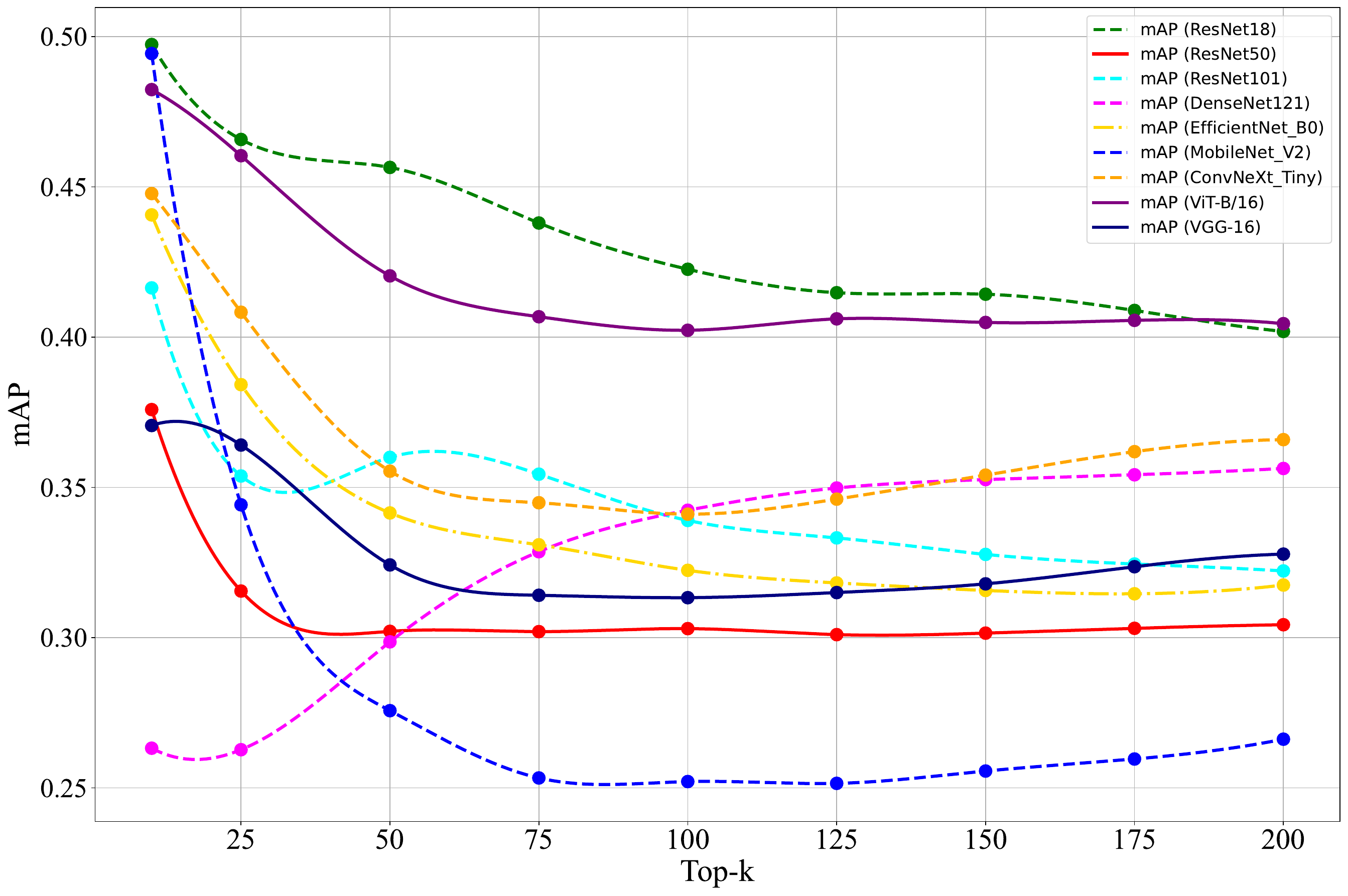}
        \caption{Comparison of various models on \textit{mean average precision} results for retrieving the top $k$ matched faces given \textbf{query skull} from the \textbf{mixed gallery face} using different models.}
        \label{fig:precision_mixed}
    \end{figure}

\begin{figure}[!t]
        \centering
        \centering
        \includegraphics[width=1.25\linewidth, keepaspectratio, trim={3.5cm 3cm -2.7cm 2cm},clip]{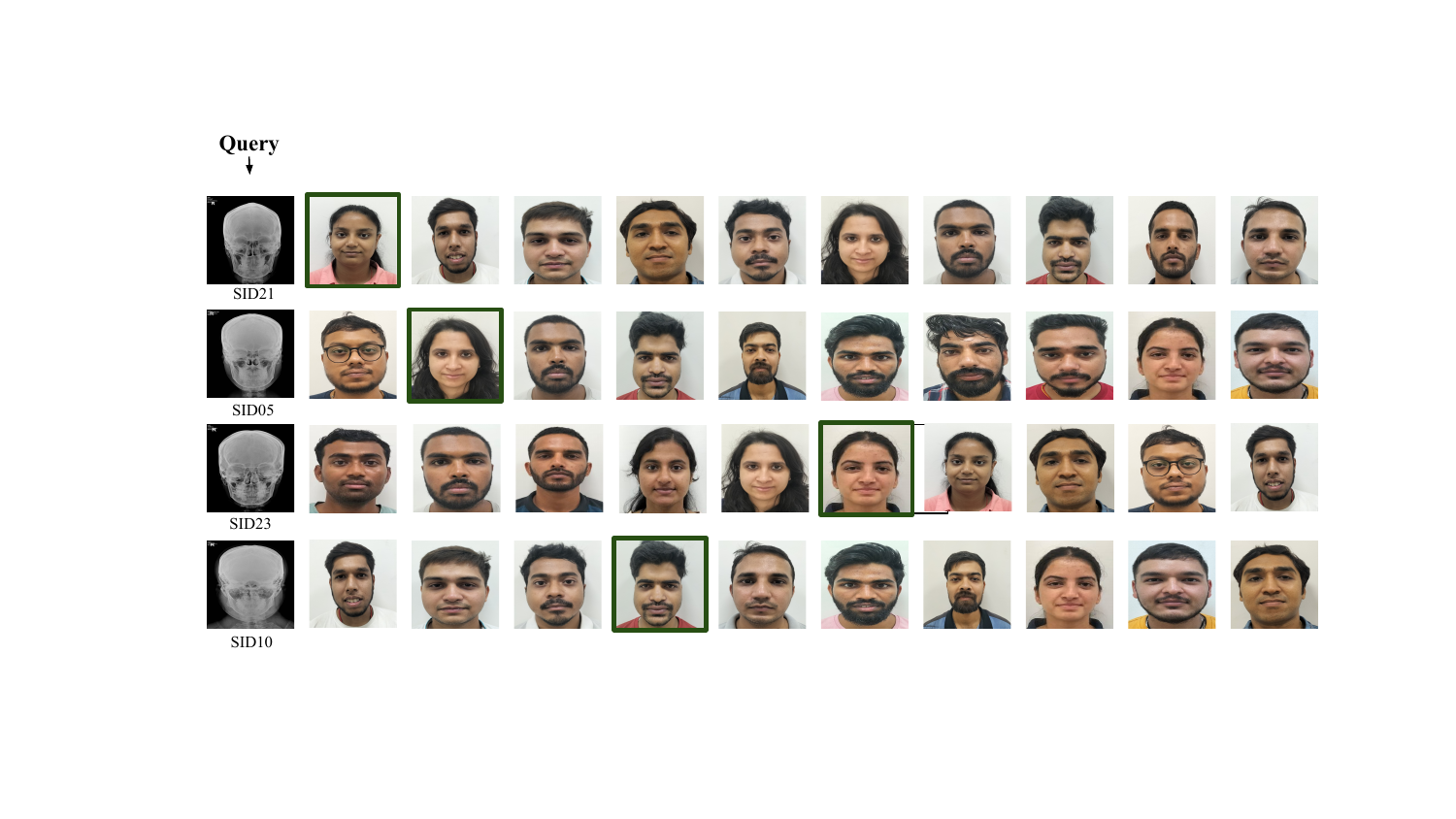}
        \textbf{\small(a) Retrieval results with ResNet50}
        \hfill
        \centering
        \includegraphics[width=1.25
        \linewidth, keepaspectratio, trim={3.5cm 3cm -2.4cm 3cm},clip]{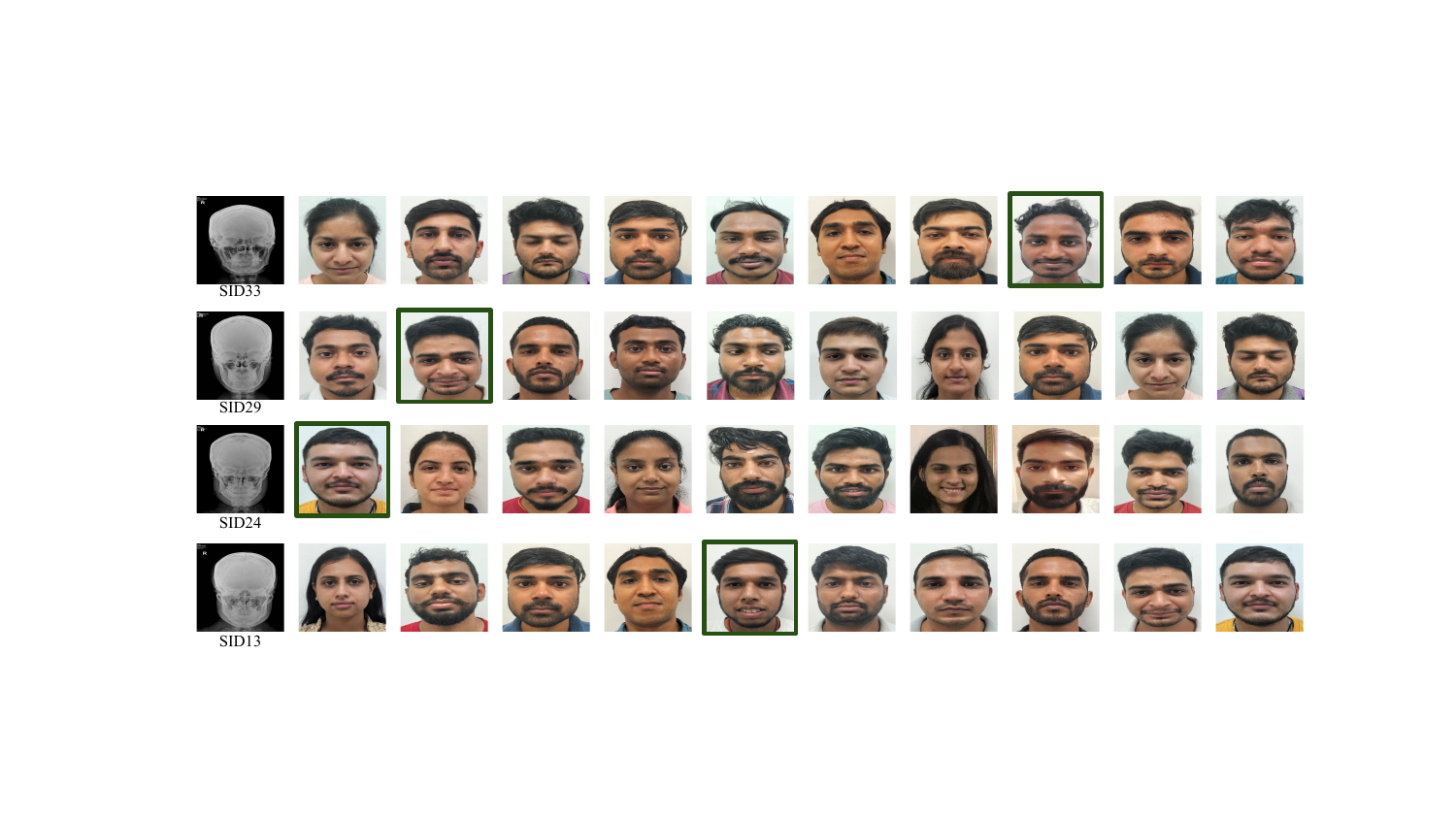}
        \textbf{\small(b) Retrieval results with DenseNet121 }
        \caption{Qualitative results of top $10$ retrievals using \textbf{(a)} ResNet50 and \textbf{(b)} DenseNet121 model. The face with a green border is the correct matched face of the given skull. $SID01$ to $SID16$ are query skulls.}
        \label{fig:retrieval}
    \end{figure}

    \begin{figure}[!ht]
        \centering
        \includegraphics[width=0.9\linewidth, keepaspectratio]{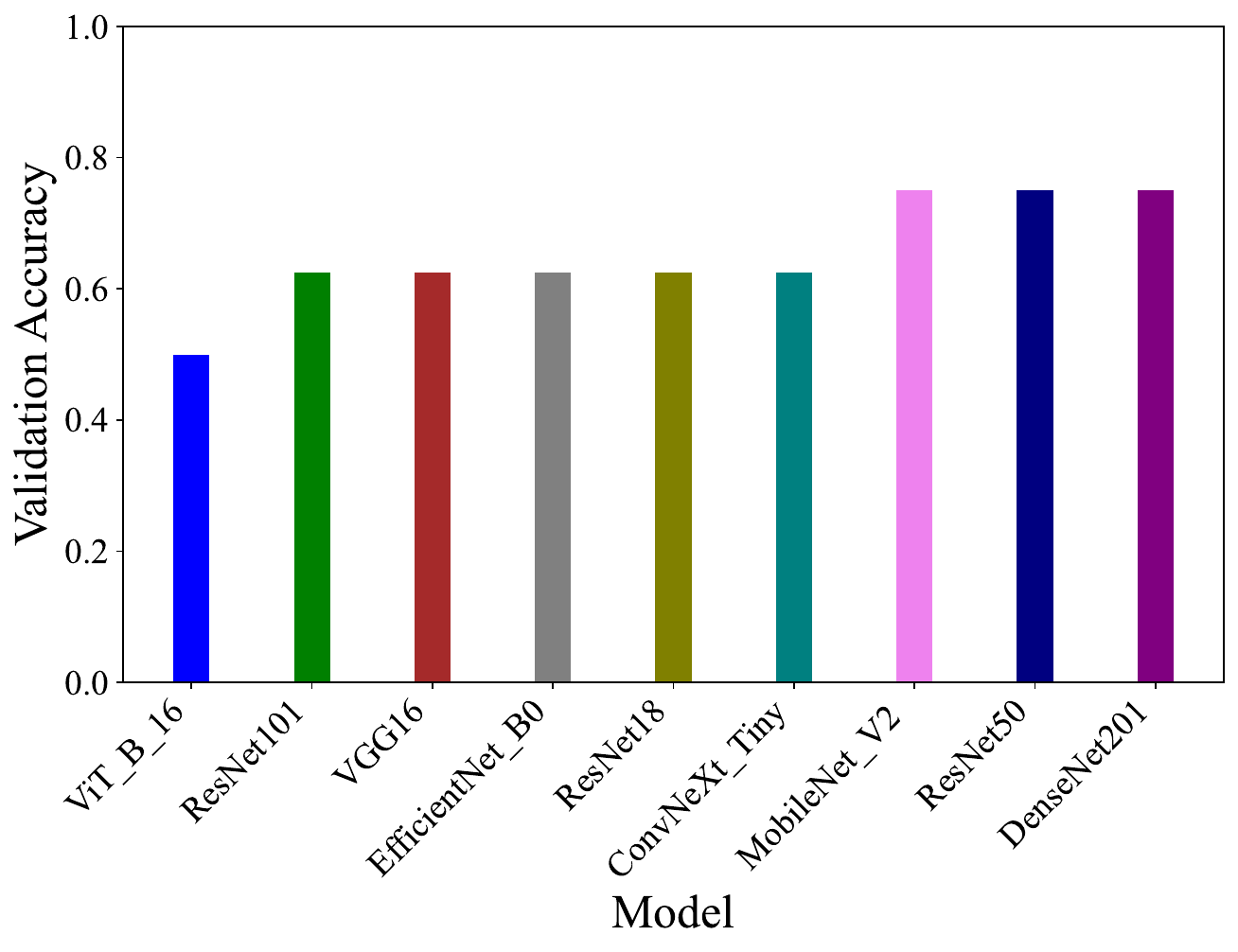}
       \caption{Comparison of different models for validation accuracy, out of which  \textbf{MobileNet\_V2}, \textbf{ResNet50}, \textbf{DenseNet201} are showing better validation accuracy. }
        \label{fig:validation_loss}
\end{figure}

The trained model thus can produce a confidence score for the given skull ($\boldsymbol{X}$) and face ($\boldsymbol{I}$) belongs to a similar person. The confidence score is computed as follows
  
\begin{equation}\label{eq:score}
    \text{Confidence Score}_{p,n} = e^{-\delta(f(x_i^a), f(x_i^{p,n}))}
\end{equation}

\begin{equation}\label{eq:distance}
        \text{where} \hspace{4mm} \delta(f(x_i^a),f(x_i^{(p,n)})) =||{f(x_i^a)-f(x_i^{(p,n)})}||^2
    \end{equation}
Here, the superscript a, p, and n indicate anchor, positive, and negative image, respectively.

 However, in the event of a natural disaster or crime, unidentified remains must be compared to a list of potential identities (\textit{i.e.} gallery of faces). To evaluate the model\textquotesingle s retrieval capabilities, top $k$ potential faces are retrieved for each skull based on their confidence score using~\eqref{eq:score} and ~\eqref{eq:distance}. 
  The overall evaluation of different models that are used as a backbone network for the proposed Siamese network is shown in Table~\ref{Table 1}.

To speed up the matching process, we used precomputed feature vectors of the gallery images, and during the search, we computed the Euclidean distance and confidence scores with the feature vector of a given skull and selected the top k different matches. Fig.~\ref{fig:3d_tsne} represents a 3D t-SNE visualization of query skull and gallery face embeddings with ground truth face embedding in which the left side of t-SNE has only 40 gallery faces without mixing with the IISCIFD dataset, and the right side has 485 gallery faces which is mixed with IISCIFD dataset\cite{katti2019you}, respectively. This visualization illustrates the distribution and correlation of embeddings from different domains in 3D space. The IISCIFD dataset was chosen because it comprises Indian faces representing both Northern and Southern regions of India. Fig.~\ref{fig:recall} and Fig.~\ref{fig:precision} show the recall and mAP curve for retrieving the top k-matched faces from the face gallery for the given query skull. To evaluate the performance of our retrieval method, we mixed our \emph{IITMandi$\_$S2F} face dataset with the IISCIFD dataset\cite{katti2019you}. We evaluated our model using the mixed face gallery, which includes a total of 485 face images. The quantitative results for retrieving the top k matched faces are shown in Fig.~\ref{fig:recall_mixed} and Fig.~\ref{fig:precision_mixed}. Some qualitative results obtained on our \emph{IITMandi\_S2F} dataset are presented in Fig.~\ref{fig:retrieval}, showing the top $10$ faces matched against the query skull in decreasing order of confidence score. The true face of the given query skull is highlighted in green. We have used a total of nine deep models, which are fine-tuned according to our benchmark dataset \emph{IITMandi$\_$S2F}. The comparison of validation accuracy between different models used as a backbone in our proposed Siamese networks is as shown in Fig.~\ref{fig:validation_loss}, out of nine models, ResNet50, DenseNet121, and MobileNet\_v2 are the best performing models.

\section{\textbf{Conclusions}}
In this paper, we have introduced a framework based on the Siamese network for skull-to-face matching of a person through cross-domain identity representation. Our method computes the similarity score between the given pair of skull and face images. If the score is higher than the threshold, it suggests that the given skull likely belongs to the corresponding person. Our experiments, conducted on cross-domain datasets, focus on how well the Siamese architecture model solve the task of face identification from a given skull image. Additionally, our novel benchmark skull-to-face datasets aim to encourage further research in skull-to-face recognition and classification tasks. Also, the prepared benchmark dataset can be used for future research on craniofacial reconstruction.
\printcredits

\subsection{\textbf{Acknowledgment}}
I want to express my sincere gratitude to the volunteers who contributed their time and effort to assist in the creation of the benchmark dataset \emph{IITMandi$\_$S2F} and also to IIT Mandi Health Center for their cooperation and assistance throughout the data collection process.

\subsection{\textbf{Data availability}}
 Our data will be made available on request.
\bibliographystyle{cas-model2-names}

\bibliography{cas-dc-sample}


\bio{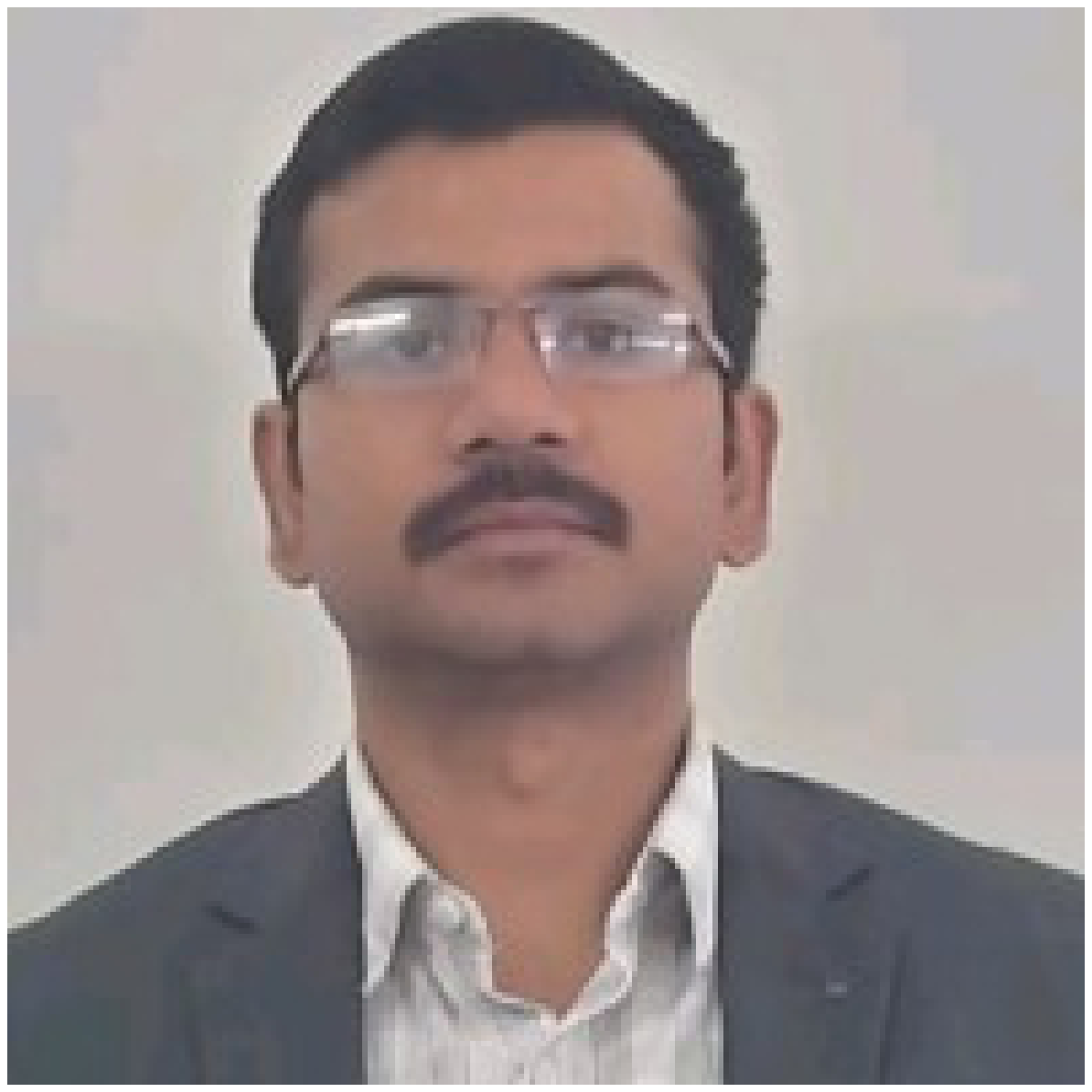}
Ravi Shankar Prasad received the BTech degree from the
Ranchi University, Jharkhand, India, in
2019 and the MTech degree in Biomedical Engineering from Jadavpur University, Kolkata, India, in 2022. He is currently working toward the
PhD degree in Intelligent Systems,
at the Indian Institute of Technology Mandi,
India. His research interests include machine learning and pattern Recognition.

\endbio

\bio{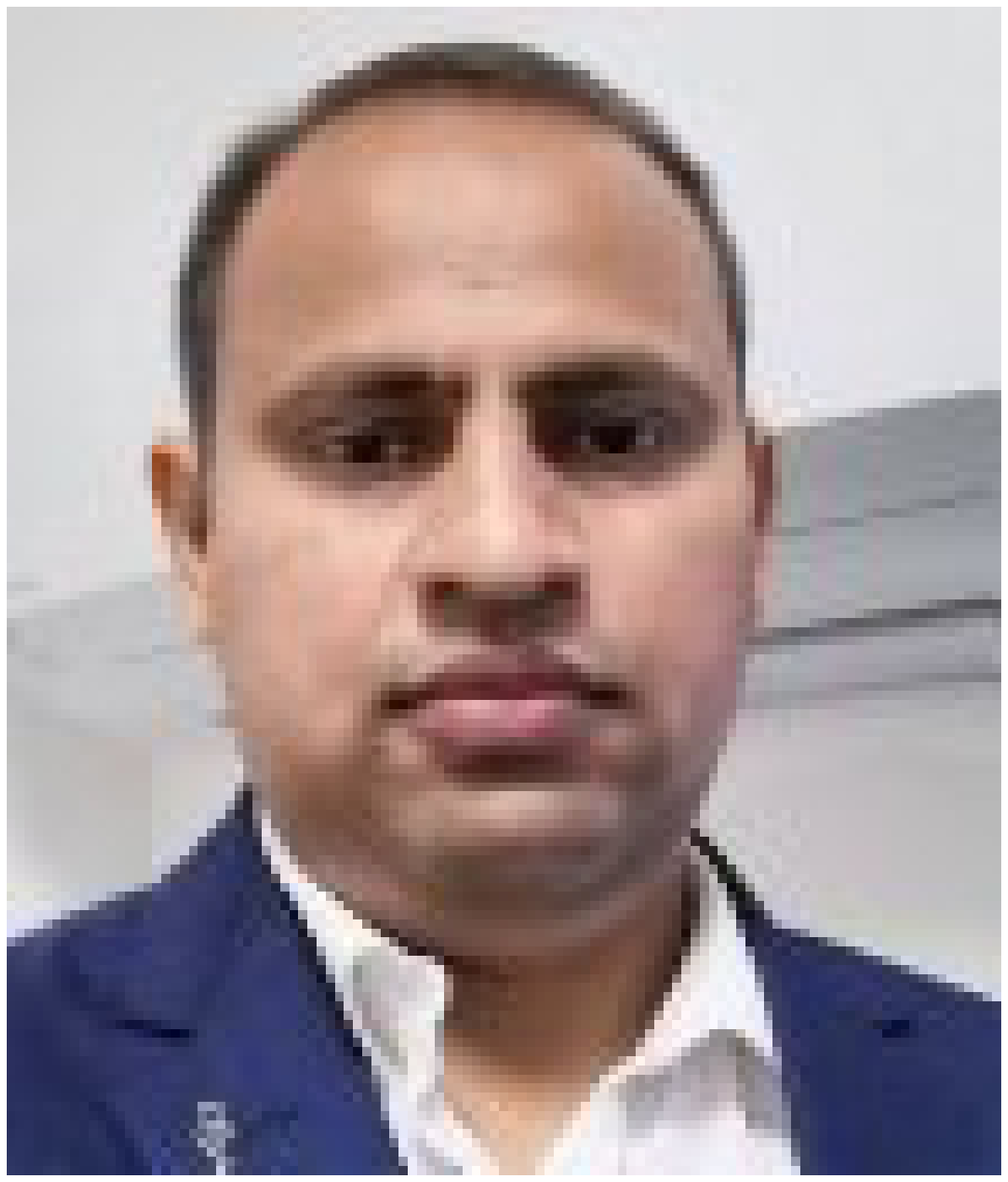}
Dinesh Singh is an Assistant Professor in the School of Computing and Electrical Engineering, Indian Institute of Technology Mandi. Prior to joining IIT Mandi, He worked in the High-dimensional Statistical Modeling Team with Prof. Makoto Yamada as a Postdoctoral Researcher at the RIKEN Center for Advanced Intelligence Project (AIP), Kyoto University Office, Japan. He completed his Ph.D. degree in Computer Science and Engineering from IIT Hyderabad on Scalable and Distributed Methods for Large-scale Visual Computing. He received his M.Tech degree in Computer Engineering from NIT Surat on machine-learning approaches for network anomaly and intrusion detection in the domain of cybersecurity and cloud security. His research interests include machine learning, big data analytics, visual computing, and cloud computing. 
\endbio

\end{document}